\documentclass{article}
\usepackage{spconf,amsmath,epsfig}
\usepackage{amsmath}
\usepackage{algorithm}
\usepackage{algorithmic}
\usepackage{color}
\usepackage{bm}
\usepackage{amssymb}
\usepackage{colortbl}
\let\OLDthebibliography\thebibliography
\renewcommand\thebibliography[1]{
  \OLDthebibliography{#1}
  \setlength{\parskip}{0pt}
  \setlength{\itemsep}{0pt plus 0.3ex}
}

\begin{document}\sloppy

\def\x{{\mathbf x}}
\def\L{{\cal L}}

%
\title{From Covert Hiding to Visual Editing: Robust Generative Video Steganography}
\name{Xueying Mao, Xiaoxiao Hu, Wanli Peng, Zhenliang Gan, Qichao Ying, Zhenxing Qian$^{\ast}$\thanks{$^{\ast}$ indicates the corresponding author. This work was supported by the National Natural Science Foundation of China under Grants U20B2051 and U1936214.}, 
Sheng Li, Xinpeng Zhang}
\address{School of Computer Science, Fudan University, China\\
\{xymao22@m., xxhu23@m., pengwanli, zlgan23@m., qcying20@, zxqian@, lisheng@, zhangxinpeng@\}\\fudan.edu.cn
}

\maketitle

\begin{abstract}
Traditional video steganography methods are based on modifying the covert space for embedding, whereas we propose an innovative approach that embeds secret message within semantic feature for steganography during the video editing process.
Although existing traditional video steganography methods display a certain level of security and embedding capacity, they lack adequate robustness  against common distortions in online social networks (OSNs).
In this paper, we introduce an end-to-end robust generative video steganography network (RoGVS), which achieves visual editing by modifying semantic feature of videos to embed secret message. We employ face-swapping scenario to showcase the visual editing effects. We first design a secret message embedding module to adaptively hide secret message into the semantic feature of videos.
Extensive experiments display that the proposed RoGVS method applied to facial video datasets demonstrate its superiority over existing video and image steganography techniques in terms of both robustness and capacity.

\end{abstract}
\begin{keywords}
Generative video steganography, Robust steganography, Semantic modification
\end{keywords}
\section{Introduction}
\label{sec:intro}

Steganography is the science and technology of embedding secret message into natural digital carriers, such as image, video, text, etc.
Generally, the natural digital carriers are called “cover” and the digital media with secret message are called “stego”.
Conventional image steganography methods \cite{you2022image,zhu2018hidden,zhang2019steganogan} primarily modify high-frequency components to embed secret message. 
They commonly utilize methodologies such as pixel value manipulation or integrating secret message into the cover image before inputting it into an encoder for steganographic purposes.

\begin{figure}[t!]
  \centering
  \includegraphics[width=1\linewidth]{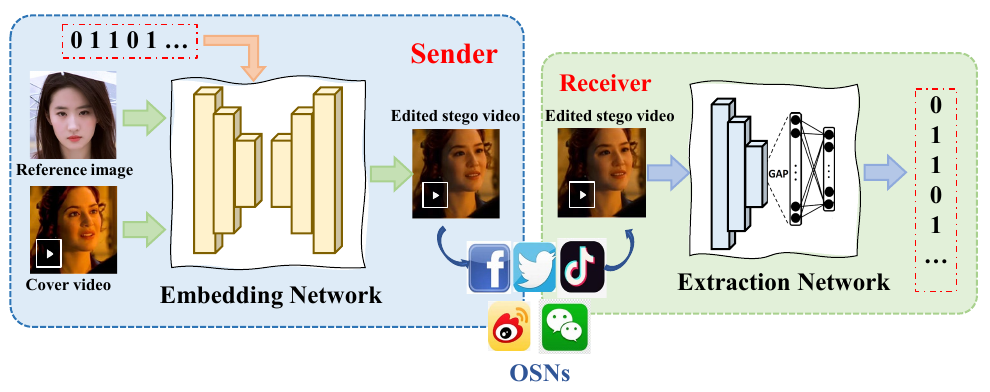}
  \caption{\textbf{Methodology of RoGVS.} We modulate semantic feature with secret message to edit videos, such as the identity feature in facial videos. Our RoGVS can generate high-quality stego videos even in the presence of various distortions.}
  \label{image_fig1}
\end{figure}

In the past few years, as the rise of short video software applications like TikTok, YouTube, Snapchat, etc., video has become a suitable carrier for steganography.
Traditional video steganographic methods, utilizing direct pixel value manipulation \cite{cetin2009new}, coding mapping \cite{liu2021high}, or adaptive distortion function \cite{he2023adaptive}, exploit video data redundancy for information hiding. Nevertheless, while successful in security and embedding capacity, these methods on modifying covert space can be erased by common post-processing operations easily. So they are vulnerable to mitigate diverse distortions that may  occur in lossy channel transmission.

\begin{figure*}[!t]
  \centering
  \includegraphics[width=1.0\linewidth]{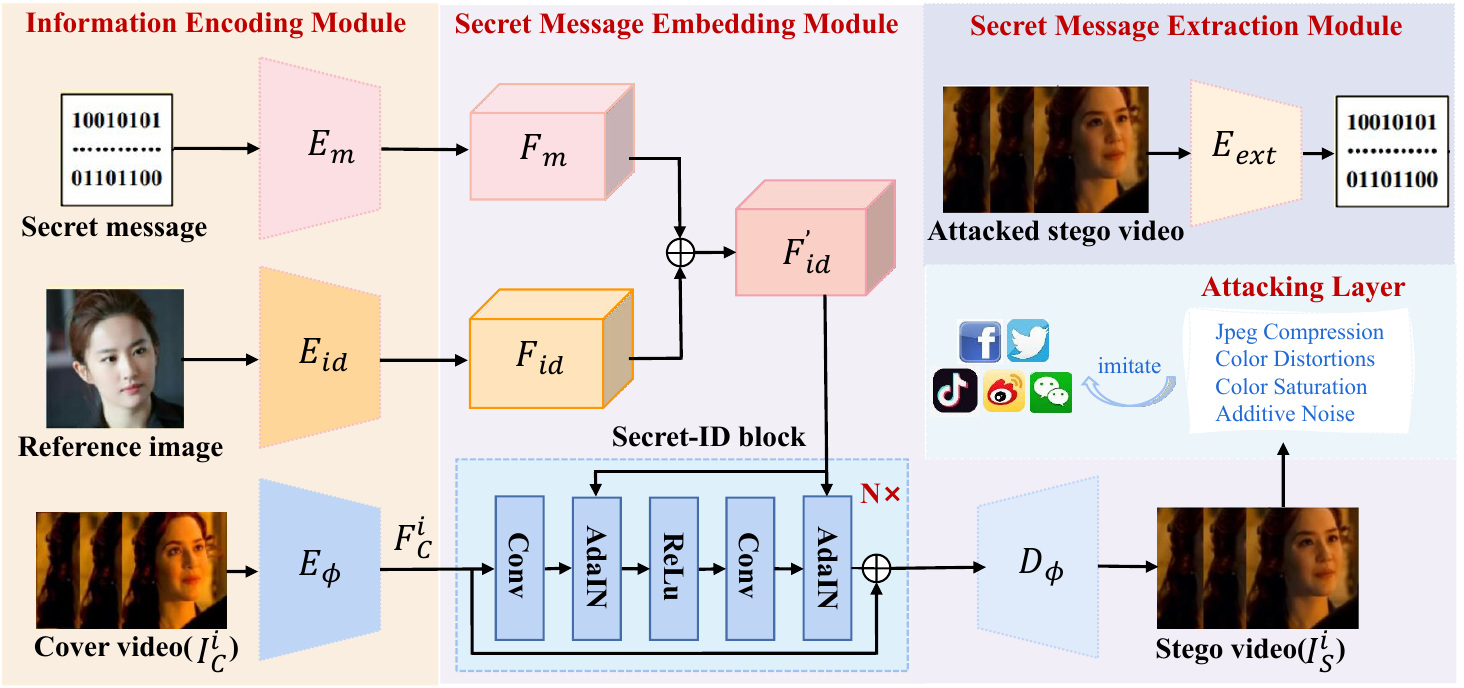}
  \caption{\textbf{The Framework of the Proposed RoGVS.} $\bm{E}_{m}$ is secret message encoder. $\bm{E}_{id}$ is identity feature extractor. $\bm{E}_{\phi}$ is video feature extractor. $\bm{D}_{\varphi}$ represents a video decoder. $\bm{E}_{ext}$ represents secret message extractor.}
  \label{fig:overall}
\end{figure*}
Visual editing on videos can be seen as the process of modifying the semantic information of objects within them.
Instead of hiding secret message in covert space, we embed secret message within semantic feature of videos for visual edition. The advanced semantic feature is less susceptible to distortions, making this method inherently robust.
In order to improve the robustness of video steganography, we propose an end-to-end robust generative video steganography network (RoGVS), which consists of four modules, containing information encoding module, secret message embedding model, attacking layer, and secret message extraction module.
For evaluation, we use face-swapping technology as an example to show the effectiveness of our method, while it can be easily extended to other applications.
Comprehensive experiments have showcased that our method surpasses state-of-the-art techniques, attaining commendable robustness and generalization capabilities.

The main contributions of our work are as follows:
1) We are the first to explore a novel generative video steganography method, which modifies semantic feature to embed secret message during visual editing instead of modify the covert space. This framework exhibits strong extensibility, serving as a new topic for the future development of the steganography field.
2) The proposed method is robust against common distortions in social network platform and the secret message can be extracted with high accuracy.
3) Our method achieves better security for anti-steganalysis than other state-of-the-art methods, which can effectively evade the detection of steganalysis system.

\section{Related Work}
\textbf{Image Steganography.}
Conventional image steganography methods primarily modify high-frequency components to embed secret message. The LSB substitution method ~\cite{chan2004lsb} operates under the assumption that human eyes cannot perceive changes in the least significant bit of pixel values.
HiDDeN ~\cite{zhu2018hidden} introduces an end-to-end trainable framework through an encoder-decoder architecture.
SteganoGAN ~\cite{zhang2019steganogan} employs dense encoders to enhance payload capacity. Wei et al ~\cite{wei2022generative} propose an advanced generative steganography network that can generate realistic stego images without using cover images.
However, alterations in high-frequency components can be obliterated by common post-processing operations, such as JPEG compression or Gaussian Blur.

\noindent\textbf{Video Steganography.} Early video steganography usually modifies RGB or YUV color spaces for embedding secret message.
Dong et al \cite{dong2019high} observed that altering intra-frame modes in HEVC significantly affected video coding efficiency, while modifications to multilevel recursive coding units had minimal distortion impact.
PWRN \cite{li2022anti} employs a super-resolution CNN, the Wide Residual-Net filter (PWRN), to replace HEVC's loop filter.
Recently, He et al \cite{he2023adaptive} devised an adaptive distortion function using enhanced Rate Distortion Optimization (RDO) and Syndrome-Trellis Code (STC) to minimize embedding distortion.
However, these methods are struggle to handle various distortions that may arise in lossy channel transmission.

\noindent\textbf{Visual Editing.}
Visual editing can encompass color correction on a single image, deletion, addition, or alteration of objects within the image, or even merging two photos to create an entirely new scene. In videos, visual editing might involve adding effects to specific frames, removing elements from the video to alter the scene, replacing one person's face with another ~\cite{chen2020simswap}, also called face-swapping.

\section{Proposed Approach}
Our method aims to embed secret message $\bm{M}$ using semantic feature extracted from reference image $\bm{I}_{R}$ into cover video $\bm{V}_C$, generating stego video $\bm{V}_{C}^{'}$. As illustrated in Fig. \ref{fig:overall}, our approach comprises four modules: Information Encoding Module, Secret Message Embedding Module, Attacking Layer, Secret Message Extraction Module.

\subsection{Information Encoding Module} 
The information encoding module consists of three parts: 
The first is identity extractor $\left(\bm{E}_{id}\right)$ which utilizes a facial recognition network to extract identity feature tailored for the reference image $\left(\bm{I}_R\right)$.
The second is video feature extractor $\left(\bm{E}_{\phi}\right)$. It acquires the latent representation of cover video $\bm{V}_{C}$ with $v$ frames, employing an encoder \cite{chen2020simswap} for video feature extraction.
The third is secret message encoder $\left(\bm{E}_m\right)$ which is a one-layer dense Multi-layer Perceptron (MLP).
The above three parts are formulated as follows:
\begin{equation}
    \bm{F}_{id} = \bm{E}_{id}(\bm{I}_{R})
\end{equation}
\begin{equation}
    \bm{F}^i_C = \bm{E}_{\phi}(\bm{I}_C^i)
\end{equation}
\begin{equation}
    \bm{F}_m = \bm{W_mM} + \bm{b}_m ,
\end{equation}
where $\bm{I}^i_C$ is the $i$-th frame of the cover video. 
$\bm{F}^i_{C}$ represents the latent feature representation of $i$-th frame. 
$\bm{F}_{id}$ is the identity feature of the reference image.
$\bm{M}$ is the secret message.
$\bm{W_m}$ and $\bm{b}_m$ represents the learnable weights and biases.

\subsection{Secret Message Embedding and Extraction Module} 
This module aims to embed the secret message during face swapping.
The key problem is how to implement face swapping under the guidance of secret message.
To our understanding, the latent features of the cover video encompass both identity and attribute feature.
Face swapping essentially involves replacing the cover video's identity with that of the reference image.
Consequently, we embed the secret message into the identity feature of the reference image, formulated as follows:
\begin{equation}
\bm{F}'_{id} = \bm{F}_{id}+\lambda \cdot \bm{F}_{m} ,
\label{eq:embed}
\end{equation}
where $\lambda$ is a hyper-parameter adjusting the influence of secret message on identity feature.

Due to strong coupling between identity and attribute features, direct extraction of attribute feature from the latent representation $\bm{F}_{c}^{i}$ by $\bm{E}_{\phi}$ is unfeasible. To ensure better attribute preservation, we design a Secret-ID block, consisting of the modified version of the residual block and AdaIN to inject $\bm{F}_{id}^{i}$ into $\bm{F}_{C}^{i}$.
The Secret-ID block is formulated as follows:
\begin{equation}
\bm{AdaIN}(\bm{F}^i_{C},\bm{F}'_{id}) = \sigma _{\bm{F}'_{id}} \frac{\bm{F}^i_{C}-\mu (\bm{F}^i_{C})}{\sigma (\bm{F}^i_{C})} +\mu _ {\bm{F}'_{id}} ,
\label{eqn_adain}
\end{equation}
where $\mu ({\bm{F}^i_{C}})$ and $\sigma ( {\bm{F}^i_{C}})$ represent the channel-wise mean and standard deviation of the input feature $\bm{F}^i_{C}$, respectively. Meanwhile, $\sigma _{\bm{F}'_{id}}$ and $\mu _{\bm{F}'_{id}}$ correspond to two variables derived from the secret-identity feature $\bm{F}'_{id}$.

After N Secret-ID blocks, the identity feature in $\bm{F}_{C}^{i}$ is replaced by $\bm{F}'_{id}$ and then we get $\bm{F}^{i}_{S}$. Subsequently, we use an video Decoder $\bm{D}_{\phi}$ to recover the $i$-th frame $\bm{I}_{S}^{i}$ of the stego video from $\bm{F}^{i}_{S}$. The Decoder contains four upsample blocks, a ReflectionPad layer and a convolutional layer. Each upsample block consists of a upsample layer, a convolutional layer and a BatchNorm layer. The process to get $\bm{I}^{i}_{S}$ can be expressed as $\bm{I}^{i}_{S} = \bm{D}_{\phi} (\bm{F}^{i}_{S})$.

We design an extraction module to retrieve secret message $\bm{M}'$ from the stego videos, featuring seven convolutional layers using ReLU activation. Ultimately, a sigmoid activation function and binarization are applied to extract the embedded secret message. This module's formulation is as $\bm{M}' = \bm{E}_{ext}\left(\bm{V}_S\right)$.

\subsection{Attacking Layer} 
To bolster the robustness of our method for face-swapping videos in real-world scenarios, we design a attacking layer. This module simulates prevalent distortions encountered across social network platforms.

\noindent\textbf{JPEG Compression.} JPEG compression involves a non-differentiable quantization step due to rounding. To mitigate this, we apply Shin et al.'s method ~\cite{shin2017jpeg} to approximate the near-zero quantization step using function Eq. (\ref{eq:JPEG}):
\begin{equation}
\begin{aligned}
\bm{q(x)} = \left \{
\begin{array}{ll}
    \bm{x^{3}},                    & \bm{|x| < 0.5}\\
    \bm{x},                           & \bm{|x|\geq 0.5}
\end{array}
\right. ,
\end{aligned}
\label{eq:JPEG}
\end{equation}
where $x$ denotes pixels of the input image.
We uniformly sample the JPEG quality from within the range of [50, 100].

\noindent\textbf{Color Distortions.} We consider two general color distortions: brightness and contrast.
We perform a linear transformation on the pixels of each channel as the formula Eq.~(\ref{eqn_bc}):

\begin{equation}
    \bm{p(x)}=a\times\bm{f(x)}  + c ,
\label{eqn_bc}
\end{equation}
where $\bm{p(x)}$ and $\bm{f(x)}$ refers to the distorted and the original image. The parameters $\bm{a}$ and $\bm{c}$ regulate contrast and brightness, respectively.

\noindent\textbf{Color Saturation.} We perform random linear interpolation between RGB and gray images equivalent to simulate the distortion.

\noindent\textbf{Additive Noise.}
We use Gaussian noise to simulate any other distortions that are not considered in the attacking layer. We employ a Gaussian noise model (sampling the standard deviation $\delta \sim U[0, 0.2])$ to simulate imaging noise.

\subsection{Loss Function} 
The proposed method ensures both high stego video quality and precise extraction of secret message. We achieve this by training the modules using the following losses.

\noindent\textbf{Identity Loss}.
The identity loss minimizes the variance between the identity features ($\bm{F}_{id}$) of the reference image and the $i$-th frame ($\hat{\bm{F}}_{id}^{i}$) in the stego video, reducing alterations caused by secret message. Cosine similarity is used to measure this loss by the formula Eq (\ref{eq:cos}).
\begin{equation}
\mathcal{L}_{\emph{id}}=1-\frac{\bm{F}_{id} \times \hat{\bm{F}}_{id}^i}{||\bm{F}_{id}||_{2}||\hat{\bm{F}}_{id}^i||_{2}} .
\label{eq:cos}
\end{equation}

\begin{table*}[h]\small
\caption{\textbf{Comparison Results on Extraction Accuracy.} ``-" means ``Without Distortion". ($\cdot$) represents Bits Per Frame (BPF). Under different distortion scenarios, our method demonstrates superior performance in comparison.}
  \label{table_comparison_results}
\renewcommand{\arraystretch}{1.2}
\centering
  \resizebox{1.0\textwidth}{!}{
\begin{tabular}{ccccccccccccccc}
\hline
{Method} &{-} &{PNG} &{Resize~(0.5)}  & {Bit Error} & {Brightness} & {Contrast}  &  {H.264 ABR} &  {H.264 CRF} &  {Motion Blur} &  {Rain} & {Saturate} & {Shot Noise}\\
\hline
HiDDeN \cite{zhu2018hidden}  & 0.9633   &  0.8342  &  0.6516  &  0.7543  &  0.7939  &  0.7813  &  0.7901  &  0.7813  &  0.7635  &  0.7624  &  0.7927  &  0.6310  \\
LSB \cite{chan2004lsb}  & \textbf{1.0000}   &  0.4988  &  0.4932  &  0.4533  &  0.4685  &  0.4985  &  0.4921  &  0.4932  &  0.4935  &  0.5085  &  0.4885  &  0.5012  \\
PWRN \cite{li2022anti} & \textbf{1.0000}   &  0.8473  &  0.6392  &  0.8082  &  0.7959  &  0.4470  &  0.7430  &  0.7907  &  0.6004  &  0.7255  &  0.7743  &  0.8291  \\
\textbf{Ours~(9)} & 0.9737    &  0.9650 &  0.8510  &  0.9393  &  0.9409  &  0.8959  &  0.8792  &  0.9566  &  0.9414  &  0.9374  &  0.9521  &  0.9059  \\
\textbf{Ours~(18)}  &  0.9942 &  \textbf{0.9665}  &  \textbf{0.9486}  &  \textbf{0.9565}  &  \textbf{0.9605}  &  \textbf{0.9544}  &  \textbf{0.9634}  &  \textbf{0.9642}  &  \textbf{0.9587}  &  \textbf{0.9623}  &  \textbf{0.9612}  &  \textbf{0.9588}  \\
\hline 
\end{tabular}}
\end{table*}

\noindent\textbf{Attribute Loss.} 
We use the weak feature matching loss ~\cite{chen2020simswap} to constrain attribute difference before and after embedding secret message.
The loss function is defined as follows:

\begin{equation}
\mathcal{L}_{att}= \sum_{j=h}^{H} \frac{1}{N_{j}}||D_{j}(\bm{I}^{i}_{S})-D_{j}(\bm{I}^{i}_{C})||_{1} ,
\label{eqn_original}
\end{equation}
where $\bm{D}_{j}$ refers to the feature extractor of Discriminator D for the j-th layer, $N_j$ is the number of elements in the j-th layer, and $H$ is the total number of layers. Additionally, $h$ represents the starting layer for computing the weak feature matching loss.

\noindent\textbf{Adversarial Loss.} To enhance performance, we use multi-scale Discriminator with gradient penalty. We adopt the Hinge version of adversarial loss defined as follows: 

\begin{equation}
\mathcal{L}_{adv}=\bm{E}_{x}[-log\bm{D}(x)]+ \bm{E}_{z}[log(1-\bm{D}(z)] ,
\label{eqn_adv}
\end{equation}
where $\bm{D}$ denotes the Discriminator, $x$ and $z$ in our method is respectively $\bm{I}_R$ and $\bm{I}_{S}^{i}$

\noindent\textbf{Secret Loss.} To address this, we use the Binary Cross-Entropy loss (BCE) as defined in Eq.~(\ref{eqn_secret}).
\begin{equation}
\mathcal{L}_{\emph{sec}}=\mathcal{L}_{\emph{bce}}(\bm{M},\bm{M^{'}}) .
\label{eqn_secret}
\end{equation}

\textbf{Total loss. }The total loss is defined as follows:
\begin{equation}
\mathcal{L}=\alpha _{1}\mathcal{L}_{\emph{id}}+\alpha _{2} \mathcal{L}_{\emph{att}}+\alpha _{3}\mathcal{L}_{\emph{sec}}+\mathcal{L}_{\emph{adv}}+\alpha _{4}\mathcal{L}_{\emph{GP}} .
\label{eqn_total_loss}
\end{equation}

\section{Experiments}

\subsection{Experimental Setups}

\noindent\textbf{Datasets.} We use Vggface2 ~\cite{cao2018vggface2} for training and FFHQ ~\cite{karras2019style} for validation. We crop and resize facial areas to a fixed 224 $\times$ 224 resolution for input images. 
To analyze quality and performance, we randomly select 100 videos from DeepFake MNIST+ ~\cite{huang2021deepfake} to evaluate the performance.

\noindent\textbf{Implementation Details.}
We train the model to encode a binary message of length $m$ = 9 or 18 bits in a frame. 
During training, we employ Adam optimizer with a learning rate of $4\times10^{-4}$ and a batch size of 4.
We set $\alpha_{1}=10$, $\alpha_{2}=10$, $\alpha_{3}=15$, and $\alpha_{4}=10^{-5}$.
The networks train for 1 million steps, integrating the Attacking Layer after the initial 800k steps for stability. We use an NVIDIA GeForce RTX 3090 GPU for our experiments.
\begin{figure}[!t]
  \centering
  \includegraphics[width=0.95\linewidth]{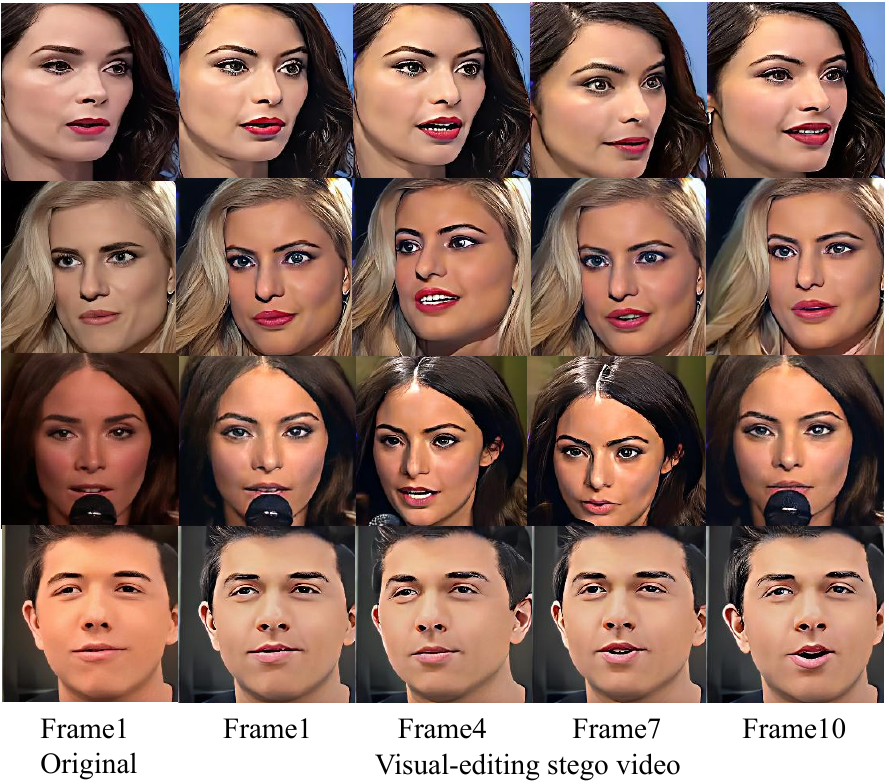}
  \caption{\textbf{Qualitative Analysis of Stego Videos. }Original represents frames within the cover videos.}
  \label{image_frame}
\end{figure}

\begin{figure*}[!t]
  \centering
  \includegraphics[width=1.0\linewidth]{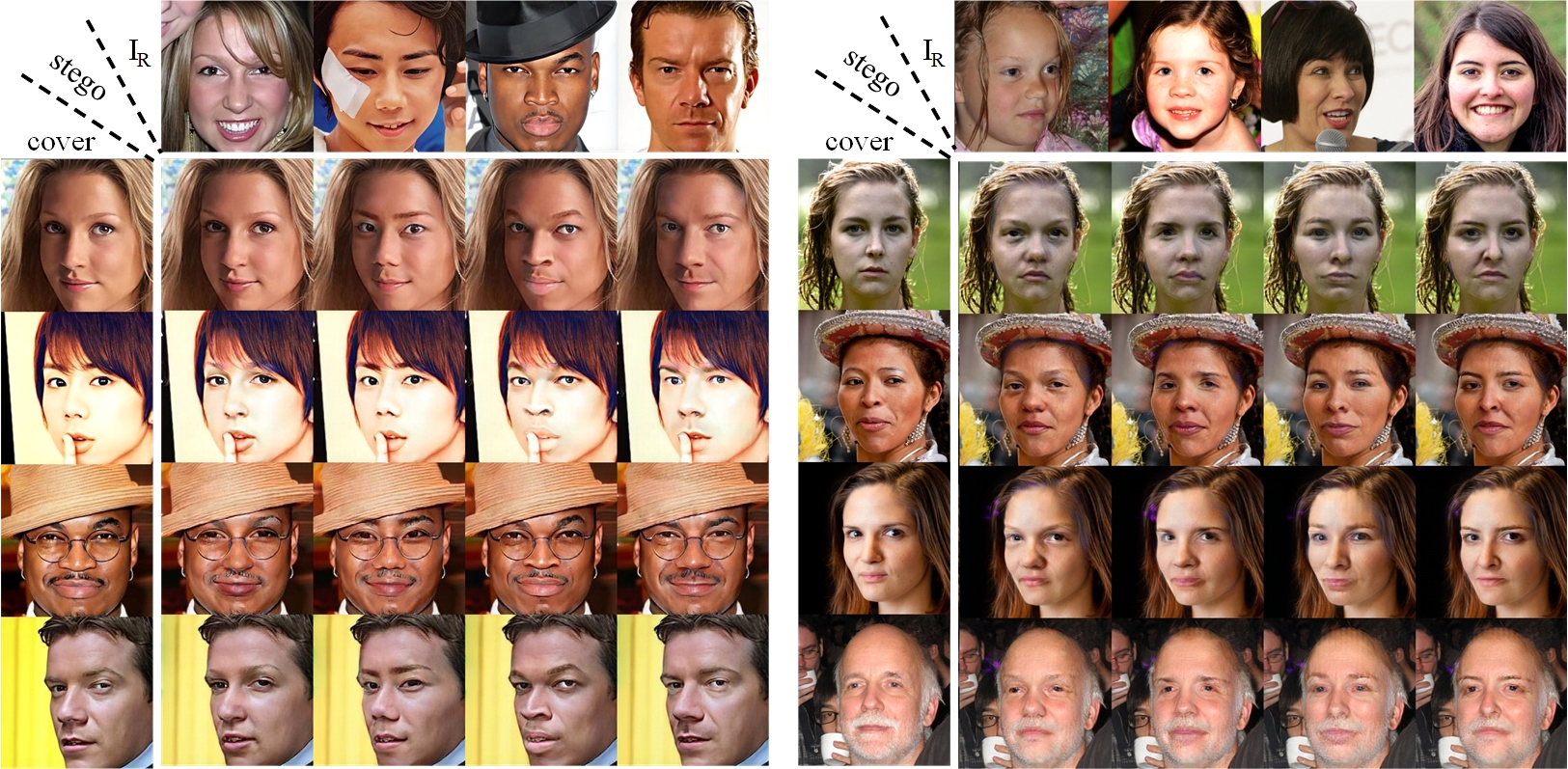}
  \caption{\textbf{Exampled Generated Stego Video Frames. }Left: Vggface2. Right: FFHQ.}
  \label{image_examples}
\end{figure*}

\noindent\textbf{Evaluation Metrics.}
We employ Bits Per Frame (BPF), quantifying the bits number of secret message per frame in the stego video.
To assess robustness, we evaluate secret message extraction accuracy under various scenarios.
For security assessment, we use three steganalysis methods \cite{zhai2019universal,li2019hevc,sheng2017prediction} to demonstrate our method's anti-detection capability.

\noindent\textbf{Baselines.}
To ensure fair comparison in our experiments, we align HiDDeN and LSB to this capacity. Detailed methods of HiDDeN and LSB are available in the supplementary materials. Additionally, due to its PU-based design, PWRN has a limited capacity of 15 BPF when resizing input images to $224\times224$.

\subsection{Performance Analysis}
We compare the performance of our RoGVS with image-level steganography including HiDDeN~\cite{zhu2018hidden} and LSB~\cite{chan2004lsb} and video-level steganography including PWRN~\cite{li2022anti}.

\noindent\textbf{Video Quality Assessment. }Fig \ref{image_examples} shows qualitative results on the integrity of generated video frames. We perform tests within and across datasets, each containing 16 test samples. The generated faces effectively change individual identities while retaining attributes like expressions and poses. More findings are available in the supplementary materials. Fig \ref{image_frame} illustrates the visual effects of certain intermittent frames within the stego videos.

\noindent\textbf{Comparisons on Extraction Accuracy \& Robustness.}
We conduct extensive experiments with multiple types of distortions.
Detailed distortion implementations are provided in the supplement.
\begin{table}[!t]
\caption{\textbf{Ablation Study on Different Embedding Positions of Secret Message.} Evaluation metric: Accuracy.}
  \label{table_ablation_position}
\renewcommand{\arraystretch}{1.2}
\centering
  \resizebox{0.5\textwidth}{!}{
\begin{tabular}{ccccccc}
\hline
{Method} &{PNG} &{Resize~(0.5)} & {H.264 CRF} &  {Motion Blur} & {Shot Noise}\\
\hline
(a)   &  0.965  &  0.918  &  0.951  &  0.941 &  0.924  \\
(b)  &  0.875  &  0.722  &  0.858  &  0.848 &  0.820  \\
(c)  &  0.939  &  0.856  &  0.894  &  0.861 &  0.893  \\
RoGVS  & \textbf{0.967}  &  \textbf{0.949}  &  \textbf{0.963}  &  \textbf{0.959} &  \textbf{0.959}  \\
\hline 
\end{tabular}}
\end{table}

\begin{table}[h]\small
\caption{\textbf{Ablation Study for $\lambda$ on Extraction Accuracy. }}
	\label{table_extraction}
\renewcommand{\arraystretch}{1.2}
	\begin{center}
	\resizebox{0.47\textwidth}{!}{\begin{tabular}{c|c|c|c|c}
		\hline
	   & $\lambda=1.00$  & $\lambda=0.1$ & $\lambda=0.01$ & $\lambda=0.005$   \\
        \hline
        Accuracy& 0.8414 & 0.5885 & \textbf{0.8607} & 0.5160    \\      
		\hline
	\end{tabular}}
	\end{center}
\end{table}

The quantitative comparison results in terms of accuracy are reported in Table \ref{table_comparison_results}.
The results show that our method can successfully extract secret message with high accuracy even after severe distortions. 
LSB ~\cite{chan2004lsb} struggles even with PNG (quantization) and HiDDeN ~\cite{zhu2018hidden}, though trained with a distortion module, can not generalize well to video-level distortions.
PWRN ~\cite{li2022anti} demonstrates robustness across numerous distortions, yet its performance remains constrained under operations such as motion blur or contrast adjustment.
The proposed RoGVS method shows superior robustness to these distortions while maintaining high extraction accuracy.
\noindent\textbf{Security Analysis.} We use three video steganalysis tools to evaluate the security of our method.
The detection performance of these three steganalysis schemes is presented in Table \ref{table_staganlysis}. Table \ref{table_staganlysis} demonstrates that our method exhibits slightly superior security compared to the three counterparts.
\begin{table}[!t]\small
\caption{\textbf{Quantitative Security Analysis. }Evaluation metric: AUC. Closer to 0.5 indicates higher performance.}
	\label{table_staganlysis}
\renewcommand{\arraystretch}{1.2}
	\begin{center}
	\resizebox{0.45\textwidth}{!}{\begin{tabular}{c|c|c|c|c}
		\hline
	 Detection method & HiDDeN & LSB & PWRN & ours\\
        \hline
        Zhai et al. ~\cite{zhai2019universal}& 0.5312 & 0.5423 & 0.5456 & \textbf{0.5245}  \\
        Li et al. ~\cite{li2019hevc} & 0.5416  & 0.5467 & 0.5411 & \textbf{0.5178}  \\
        Sheng et al. ~\cite{sheng2017prediction} & 0.5309 & 0.5189 & 0.5167 & \textbf{0.5146} \\
		\hline
	\end{tabular}}
	\end{center}
\end{table}

\subsection{Ablation Study}
\noindent\textbf{Embedding Position of Secret Message.}
In our generation network with 9 Secret-ID blocks, we explore different positions for embedding the secret message. We divide the secret message into two 9-bit segments and allocate their positions.
In detail, Setting (a): 1st-4th blocks and 5th-9th blocks. 
Setting (b): 1st-2nd blocks and 3rd-4th blocks.
Setting (c): 5th-6th blocks and 7th-8th blocks. 
They are in comparison of the standard setting of RoGVS: 1st-3rd blocks and 4th-6th blocks.

Table \ref{table_ablation_position} displays the performance for these four setups. Both Settings b and c show a considerable decrease compared to Settings a and d, suggesting that adding more Secret-ID blocks improves performance. Notably, Setting c outperforms Setting b, indicating the higher influence of subsequent blocks on the generated image.

\begin{figure}[!t]
  \centering
  \includegraphics[width=0.8\linewidth]{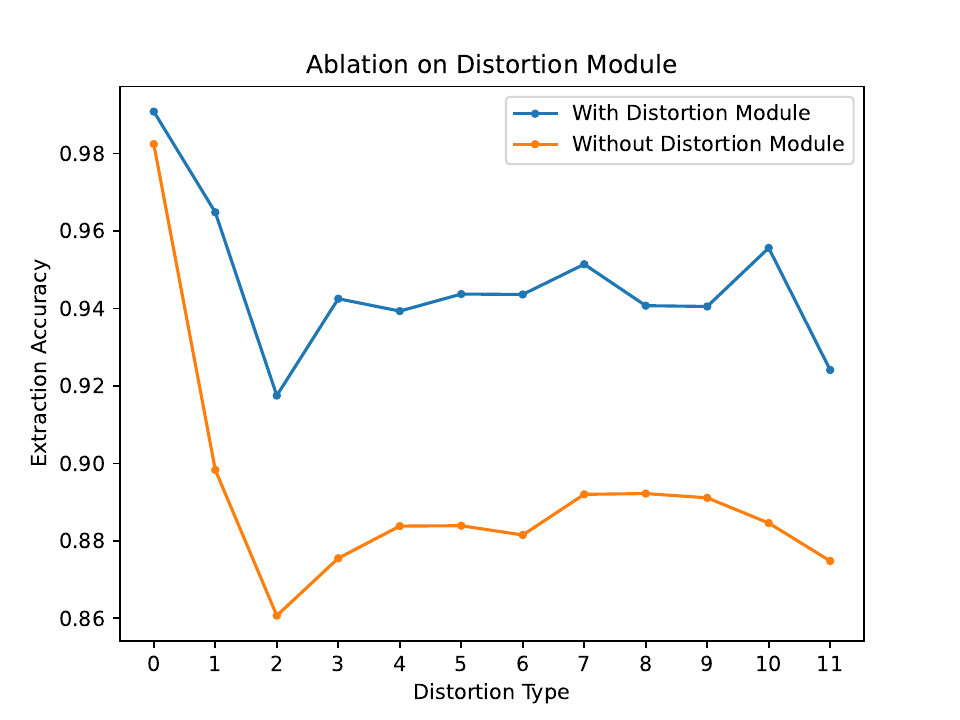}
  \caption{\textbf{Ablation Results on Attacking Layer.} The horizontal axis represents distortion types, corresponding to the order listed in Table \ref{table_comparison_results}.}
  \label{image_ablation}
\end{figure}

\noindent\textbf{Ablation on Attacking Layer, $\lambda$ \& Discriminator.}
Fig \ref{image_ablation} shows even without the module, our method demonstrates considerable robustness, surpassing the three comparative methods. The addition of attacking layer improves accuracy by an average of 6\%.
Table \ref{table_extraction} presents the impact of $\lambda$ on the extraction accuracy. More ablation results on $\lambda$ and the discriminator are displayed in the supplement. 

\section{Conclusions}
We propose a robust generative video steganography method based on visual editing, which modifies semantic feature to embed secret message. 
We use face-swapping scenario as an example to show the effectiveness of our RoGVS. The results showcase that our method can generate high-quality visually edited stego videos. What's more, RoGVS outperforms existing video and image steganography methods in robustness and capacity.

\bibliographystyle{IEEEbib}
\small
\bibliography{icme2023template}

\end{document}